\documentclass[conference]{ieeeconf}
\usepackage{amsmath,amssymb,amsfonts}
\usepackage{algorithmic}
\usepackage{graphicx}
\usepackage{textcomp}
\usepackage{booktabs}
\usepackage{xcolor}
\usepackage{soul}
\usepackage[final]{changes}
\usepackage{multirow}
\usepackage{tablefootnote}
\IEEEoverridecommandlockouts
\usepackage[hidelinks]{hyperref}

\DeclareMathOperator*{\argmin}{argmin}

\title{Learning Augmented, Multi-Robot Long-Horizon Navigation in Partially Mapped Environments}

\author{Abhish Khanal and Gregory J. Stein%
\thanks{Abhish Khanal and
        Gregory J. Stein are with the Department of Computer Science, 
        George Mason University,
        Fairfax, VA, 22030, USA.
        {\tt\small \{akhanal7, gjstein\}@gmu.edu}}%
}

\begin{document}

\maketitle

\thispagestyle{empty}
\pagestyle{empty}

\begin{abstract}
We present a novel approach for efficient and reliable goal-directed long-horizon navigation for a multi-robot team in a structured, unknown environment by predicting statistics of unknown space. Building on recent work in learning-augmented model based planning under uncertainty, we introduce a high-level state and action abstraction that lets us approximate the challenging Dec-POMDP into a tractable stochastic MDP. Our Multi-Robot Learning over Subgoals Planner (MR-LSP) guides agents towards coordinated exploration of regions more likely to reach the unseen goal. We demonstrate improvement in cost against other multi-robot strategies; in simulated office-like environments, we show that our approach saves 13.29\% (2 robot) and 4.6\% (3 robot) average cost versus standard non-learned optimistic planning and a learning-informed baseline.
\end{abstract}

\section{Introduction}

We aim to navigate through an unknown environment using multiple robots to find an unseen point goal in minimum expected distance: e.g., for package retrieval. To perform well, the multi-robot team needs to collectively navigate the unexplored region, seeking out promising routes to the goal while avoiding regions that typically lead to dead-ends.

Planning well in an unknown environment requires making inferences about unseen parts of the environment; in theory, robots must envision all possible configurations of the unknown space---including regions unlikely to lead to the goal---to determine how to navigate so that they can most quickly reach the unseen goal.
Multi-robot planning under uncertainty can be modeled as a Decentralized Partially Observable Markov Decision Processes (Dec-POMDP) \cite{DECPOMDP_MAIN1,DECPOMDPMain2}. However, Dec-POMDP planning is computationally intractable in general~\cite{DecPOMDPintractable} and it requires access to a distribution over possible environments, difficult to obtain in general.

Learning is often used to help make inferences about unseen space needed to inform good behavior and is an increasingly powerful tool for planning under uncertainty~\cite{gupta,richter1} and Dec-POMDP planning \cite{DecPOMDPusingLearning}.
However, despite impressive progress in this domain, particularly for model-free approaches trained via deep reinforcement learning~\cite{marl, rl1, rl2}, many such strategies struggle to learn effectively in large-scale environments and can be brittle in practice~\cite{RLsurvey}.

\begin{figure}[t]
        \centering
        \includegraphics[width=8.45cm]{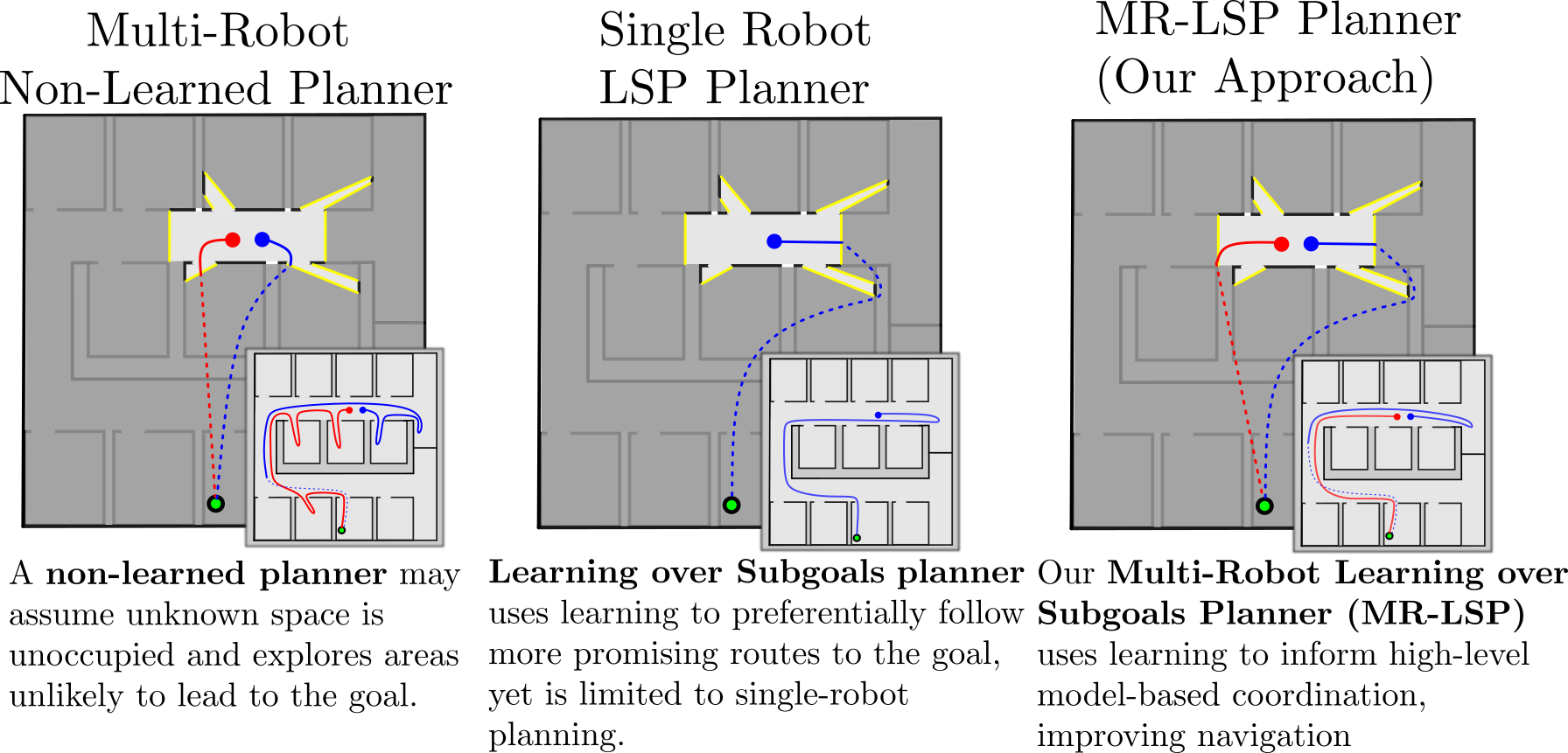}
        \caption{Our \textbf{Multi-Robot Learning over Subgoals Planner (MR-LSP)} uses learning to guide robots toward coordinated exploration of regions more likely to reach the unseen goal.}
        \vspace{-2em}\label{fig:intro}
\end{figure}
So as to avoid the computational expense of planning over the space of short-time-horizon primitive actions, many approaches for multi-robot planning introduce a topological action abstraction that simplify planning~\cite{frontier-multirobot,topological-mr,frontier-mr-1}. Such approaches typically constrain robot motion so that each robot intends to leave known space through different \emph{frontiers}, each a boundary between free and unseen space, or via paths constrained to belong to different relative homology \cite{bhattacharyaTopologicalConstraints, BhattacharyaHomologyPathPlanning}.
Though these action abstractions simplify planning and coordination between robots, action selection typically relies on simple, greedy heuristics to decide where each robot should navigate next and do not alleviate the challenges of predicting the impact of an action.

The recent Learning over Subgoals Planning abstraction (LSP)~\cite{stein2018learning} overcomes this limitation for single-robot planning.
In LSP, learning is used to estimate the goodness of high-level (frontier-associated) actions that enter unseen space, including both the likelihood that such an action will reach the goal and its expected cost.
Learning augments a model-based planning abstraction, affording performant and reliable navigation under uncertainty.
However, despite LSP's improved performance in single-robot planning under uncertainty, the LSP state transition model is not straightforwardly extended to support multiple robots, which must have the capacity to concurrently explore different unseen regions.



Leveraging insights from multi-robot topologically constrained planning and the recent Learning over Subgoals Planning (LSP) abstraction, we introduce a multi-robot generalization of the LSP model: Multi-Robot Learning over Subgoal Planning (MR-LSP). Our approach supports multi-robot planning, is reliable, and leverages learning to inform where robots should navigate next.
Each robot's high-level (topologically-constrained) actions correspond to navigation to a \emph{subgoal}---associated with a frontier---and then navigation beyond in an effort to reach the goal.
We introduce a new state abstraction and transition model that allows our high-level planner to envision how the robot team will redistribute effort once each robot finishes their respective exploratory action, a key feature of coordinated multi-robot planning.
Our abstraction lets us approximate the challenging Dec-POMDP as a stochastic MDP and solve it via sample-based tree search.
Learning is used to estimate the goodness of each action and informs planning via a Bellman Equation for our model-based planning abstraction.

We demonstrate the effectiveness of our approach in a simulated office floorplan environment, showing that our approach reduces cost by 13.29\% (two robots) and 4.6\% (three robots) versus standard non-learned optimistic planning and a competitive learning-informed baseline of our own design.


\section{Problem Formulation}
Our multi-robot team is placed in a partially-mapped environment and tasked to find a point-goal located in unseen space in minimum expected cost (distance).
Each robot is equipped with a planar laser scanner, which it uses to localize and build an accurate map of its local surroundings, a problem setting recently coined by Merlin et al.~\cite{merlin2020locally} as a \emph{Locally} Observable Markov Decision Process (LOMDP).
We assume lossless communication between robots so that all robots have an up-to-date partial map represented as an occupancy grid, as revealed by all members of the team and the poses of each robot---collectively, the team's belief $b_t$.
Planning is centrally coordinated, and so at each time step, the team's collective action $a_t$ specifies how each robot should move so as to make progress towards the unseen goal in an effort to minimize the expected cost.

Formally, we represent this problem as a Partially Observable Markov Decision Process \cite{kaelbling1998planning,littman1995learning} (POMDP). The expected cost $Q$ under this model can be written via a belief space variant of the Bellman equation \cite{Pineau-2002-8519}:
\begin{multline}\label{eq:POMDP}
    Q(b_t,a_t) = \sum_{b_{t+1}} P(b_{t+1}|b_t,a_t)\big[R(b_{t+1},b_t,a_t) \\[-10pt] + \min_{a_{t+1} \in \mathcal{A}(b_t+1)}Q(b_{t+1},a_{t+1})\big]
\end{multline}
Where $R(b_{t+1},b_t,a_t)$ is the cost accumulated by reaching belief state $b_{t+1}$ from $b_t$ by taking action $a_t$.
\begin{figure*}[!ht] 
  \vspace{0.6em}
  \includegraphics[width=0.98\textwidth]{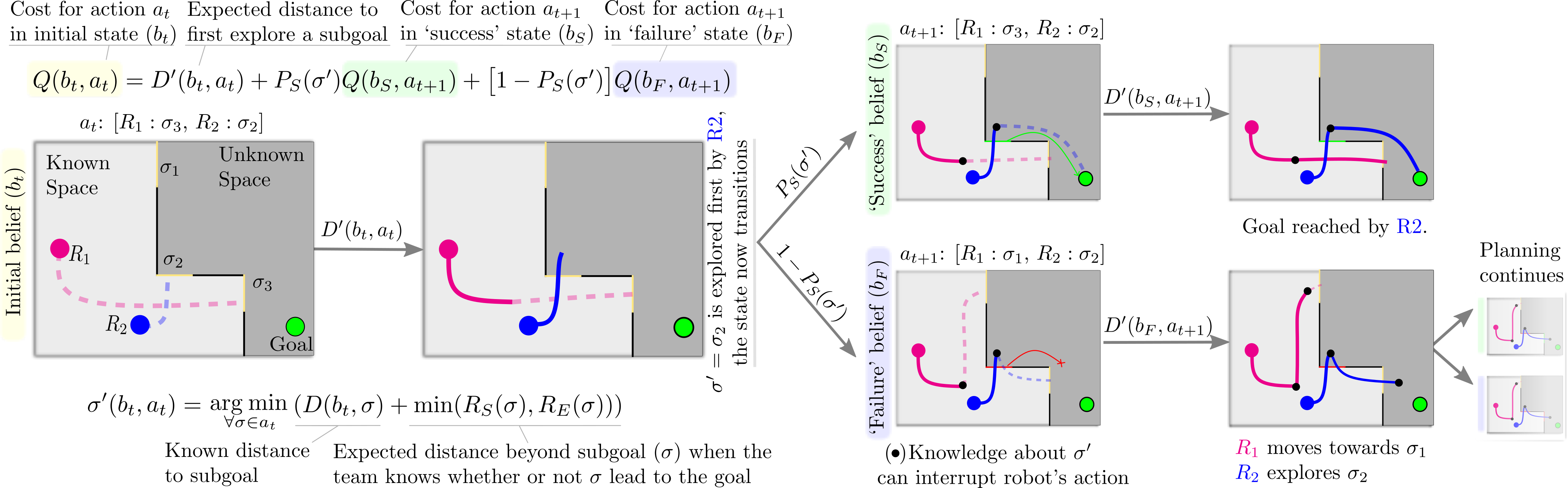}
  \caption{\textbf{Schematic of MR-LSP multi-robot expected cost calculation}\quad{}A schematic showing how the cost of collective-action under a policy is calculated using the Multi-Robot Learning over Subgoals (MR-LSP) abstraction. For action $a_t$, the robot-team concurrently explores multiple subgoals until one of the robot discovers whether its assigned subgoal leads to the goal, leading to a transition in the abstract belief state.  \vspace{-1.5em}}\label{fig:mrlsp-schematic}

\end{figure*}

\section{Preliminaries: Single-Robot Planning in a Partial Map via Learning over Subgoals} \label{sec:LSP}
Even for single robot planning, planning via the POMDP model in Eq.~\eqref{eq:POMDP} requires both enormous computational effort and also access to a distribution over possible environments, difficult to obtain in general.
So as to mitigate the complexities of \emph{single-robot} navigation in a partial map, the \emph{Learning over Subgoals Planner} (LSP) \cite{stein2018learning} introduces an abstraction in which actions correspond to navigation through \emph{subgoals} placed at boundaries between free and unknown space, simplifying the process of imagining the impact of actions that enter unseen space.
Under the LSP abstraction, model-based planning is augmented via predictions from learning, allowing for both reliability and good performance.

For the \replaced{LSP}{Learning over Subgoals planning} abstraction, temporally-extended actions correspond to \emph{subgoals}, each associated with a contiguous boundary between free and unseen space.
A \emph{high-level action} $a_t$ consists of (1) navigating to the subgoal and then (2) exploring the unknown space beyond in an effort to reach the unseen goal.
Planning is done over an abstract belief state: a tuple $b_t = \left<m_t, \mathcal{S}_u, q_t\right>$, where $m_t$ is the (partial) map of the environment, $\mathcal{S}_u$ is the set of unexplored subgoals, and $q_t$ is the robot pose. Each high-level action $a_t \in \mathcal{S}_u$ has a binary outcome: with probability $P_S(a_t)$, the robot \emph{succeeds} in reaching the goal or (with probability $1 - P_S(a_t)$) \emph{fails} to reach the goal.

Upon selecting an action $a_t$, the robot must first move through known space to the boundary, accumulating a cost $D(m_t, q_t, a_t)$.
If the robot succeeds in reaching the goal, it accumulates a \emph{success cost} $R_S(a_t)$, the expected cost for robot to reach the goal, and navigation is complete.
If it fails to reach the goal, the robot accumulates a cost associated with exploring the region and needing to turn back $R_E(a_t)$ and the state is updated to reflect that the robot has moved (to position $q(a_t)$) and that the subgoal associated with $a_t$ is explored: $b_{t+1} = \left<m_t, \mathcal{S}_u\backslash \{a_t\}, q(a_t)\right>$.
Upon failing to reach the goal, the robot must subsequently select another action from the set of unexplored subgoals: $a_{t+1}\in\mathcal{S}_u\backslash \{a_t\}$.

The expected cost of a high-level action $a_t$ can be written as a Bellman Equation:\footnote{The statistics of belief ($P_S$, $R_S$, and $R_E$) are used as input to the planner for planning, instead of the belief $b_t$, in order to transform the challenging POMDP in Eq.~\eqref{eq:POMDP} into a simpler stochastic MDP in Eq.~\eqref{eq:LSP}.}
\begin{multline}\label{eq:LSP}
    Q(b_t,a_t \in \mathcal{S}_u) = D(m_t, q_t, a_t) + P_S(a_t)  R_S(a_t) + \\\quad{}
    (1- P_S(a_t))\bigg[ R_E(a_t) + \!\! \min_{a_{t+1}\in\mathcal{S}_u\backslash \{a_t\}} Q(b_{t+1},a_{t+1})) \bigg]
\end{multline}

The terms $P_S$, $R_S$, and $R_E$---too difficult to compute exactly---are estimated from images collected on board the robot via learning.


\section{Multi-Robot Learning over Subgoals Planning (MR-LSP)} \label{sec:MR-LSP}
For single robot planning, the Learning over Subgoals planning paradigm (LSP) has demonstrated state-of-the-art performance under uncertainty and reliability-by-design, despite its reliance on learning.
Here, we extend the LSP state and action abstraction to support multi-robot planning.

While the LSP action abstraction is designed for single robot planning, generalizing the space of high-level actions to incorporate multi-robot planning is straightforward---the \emph{collective} high-level action assigns each robot a subgoal for it to navigate towards and explore beyond.
As such the collective action for an $N$-robot team can be written as a list of subgoals: $a_t = [\sigma_1, \sigma_2, \cdots, \sigma_N]$.

However, generalizing the LSP model to support multiple robots is made complicated by the fact that different robots are executing their respective subgoal-actions \emph{concurrently} and therefore may finish exploration at different times; when one robot has finished exploration beyond one subgoal, the others may not yet be done and may not even have reached the unseen space they seek to explore.
Moreover, whenever one robot completes a subgoal-action, the planner has the capacity to reassign the actions of each robot and thus where they should travel next.
If planning is to take into account the impact of concurrent action execution, we must augment the LSP state abstraction and state transition model to incorporate these effects.

In this section, we introduce our Multi-Robot Learning over Subgoals Planner (MR-LSP), which introduces a new state and state transition abstraction to model concurrent action execution, an essential component of multi-robot coordination.
Key to our approach is the recognition that estimates of the expected time to complete an action (the costs of success $R_S$ and failure $R_E$) tell us which robot is expected to complete its exploration first, knowledge we can use to simplify the process of imagining the future during planning.
We introduce a Bellman Equation for our new multi-robot planning abstraction (Sec.~\ref{sec:mrlsp:detail}), which we use in combination with a Monte-Carlo Tree Search-based approach to compute the expected cost of multi-robot collective action (Sec.~\ref{sec:pouct}).

\subsection{Expected Cost of Multi-Robot High-Level Actions}\label{sec:mrlsp:detail}
Under the MR-LSP model, the abstract state of the environment is a tuple $b_t = \left<m_t, \mathcal{S}_u, \mathcal{S}_g, q_t\right>$,  where $m_t$ is the map, $\mathcal{S}_u$ is the set of unexplored subgoals, $\mathcal{S}_g$ is the set of subgoals that are known to lead to the goal and $q_t$ is a list of the robot poses.
A high-level \emph{collective action} $a_t$ assigns each robot a subgoal to explore.
For an abstract state $b_t$, the set of high-level actions $\mathcal{A}(b_t)$ specifies all collective-actions for the robot team: $\mathcal{A}(b_t) =\bigotimes_{i\in I}(\mathcal{S}_u \cup \mathcal{S}_g)$,  constrained such that no two robots can explore same subgoal whenever possible.
A collective-action $a_t$ is thus a list of subgoals, that specifies to which subgoal each robot will next navigate to and explore beyond in an effort to reveal the goal.

Key to defining our state transition model is the idea that the multi-robot team can select a new collective-action whenever a \emph{single} robot reveals whether or not a subgoal will lead to the goal; the high-level state is updated when the \emph{first} of the robot's subgoal-actions is completed.
As in the LSP model, under our multi-robot abstraction we estimate how long it will take a single robot to either reach the goal ($D + R_S$) or to explore ($D + R_E$) beyond a particular subgoal $s$.
Given estimates of the costs of success $R_S$ and exploration $R_E$, we can determine which of the robots is expected to complete its high-level action first, yielding both (i) the subgoal $\sigma'$ that the team will first discover either does or does not lead to the goal and (ii) how long this discovery will take $D'$. Mathematically,
\begin{equation}\label{eq:frontiers-and-cost}
\begin{split}
D'(b_t, a_t) &= \min_{\forall \sigma \in a_t}\!\left(D(b_t, \sigma) + \min(R_S(\sigma), R_E(\sigma))\right) \\
\sigma'(b_t, a_t) &= \argmin_{\forall \sigma \in a_t}\!\left(D(b_t, \sigma) + \min(R_S(\sigma), R_E(\sigma))\right) 
\end{split}
\end{equation}

Under our model, the outcome of a collective action $a_t$ reveals only whether $\sigma'$ leads to the goal or not, and so the outcome of that high-level collective action is binary.
After the execution of collective action ($a_t$),
with probability $P_S(\sigma')$ the state transitions to an abstract \emph{success state} ($b_{S}$) where the robots can reach goal from subgoal $\sigma'$ or with probability $1-P_S(\sigma')$ the state transitions to an abstract \emph{failure} state ($b_{F}$) where the robots cannot reach the goal from subgoal $\sigma'$. 
Under both success and failure, the subgoal $\sigma'$ becomes explored, and so is removed from $\mathcal{S}_u$, and the robot have traveled a distance $D'$ towards completing their respective high-level subgoal-actions, ending up at new poses $q_t(a_t, D')$.
The tuple representation of abstract states associated with success ($b_{S}$) and failure ($b_{F}$) to find a path to the goal is:
\begin{align}\label{eq:updated-beliefs}
    b_{S} &= \left<m_t, \mathcal{S}_u' =  \mathcal{S}_u \backslash \{\sigma'\}, \mathcal{S}_g' = \mathcal{S}_g \cup \{\sigma'\}, q_t(a_t, D')\right> \nonumber \\
    b_{F} &= \left<m_t, \mathcal{S}_u' =  \mathcal{S}_u \backslash \{\sigma'\}, \mathcal{S}_g' = \mathcal{S}_g, q_t(a_t, D')\right>
\end{align}

Given our state and action abstraction and state transition model, we can write a Bellman Equation that defines the expected cost $Q$ for a collective action $a_t$ (see also Fig.~\ref{fig:mrlsp-schematic}):
\begin{multline} \label{eq:MR-LSP}
    Q(b_t, a_t \in \mathcal{A}(b_t)) = D' + P_S(\sigma')  \min_{a_{t+1} \in \mathcal{A}(b_{S})} Q(b_{S},a_{t+1}) \\
    +\big[1 - P_S(\sigma')\big] \min_{a_{t+1} \in \mathcal{A}(b_{F})}Q(b_{F},a_{t+1})
\end{multline}
Difficult to compute exactly, the terms $P_S$, $R_S$, and $R_E$ are estimated via learning from images collected on the robot; see Sec.~\ref{sec:Training} for a discussion of our neural network structure and training process.
Owing to the combinatorial explosion of possible actions, we cannot tractably plan while including all subgoals in the set of candidate actions. Following the example of LSP~\cite{stein2018learning}, we limit the number of subgoals under consideration to seven for all MR-LSP experiments.

We note that when a single robot is used, our MR-LSP equation reduces to the single-robot LSP model of Stein et al.~\cite{stein2018learning}, and so our approach extends the original LSP model to support multi-robot planning.\footnote{Owing to a change in how exploration is treated during concurrent action execution in our MR-LSP model---see Eq.~\eqref{eq:frontiers-and-cost}---the single-robot exploration cost $R_E$ of Eq.~\eqref{eq:LSP} becomes $R_E \leftarrow \text{min}(R_S, R_E)$, a slight deviation from the original LSP definition. With this (small) change, one-robot MR-LSP via Eq.~\eqref{eq:MR-LSP} is equivalent to single-robot LSP planning via Eq.~\eqref{eq:LSP}.}

\subsection{Navigation via MR-LSP}\label{sec:mrlsp:navigation}
The high-level collective action defines the long-horizon robot behavior.
Upon selecting a collective action $a_t$ via Eq.~\eqref{eq:MR-LSP}, each robot makes progress towards each subgoal.
We use an A$^{\!*}$ plan cost computed over the observed grid to select motion primitives for each robot that make progress towards their assigned subgoal.
The robots (i) move according to these primitive actions, (ii) observe their surroundings, (iii) update the partial map and the set of subgoals, and (iv) compute a new collective action based on this newly-updated map.
This process repeats until the goal is reached.

\section{Computing MR-LSP Expected Cost Via Sample-Based Tree Search}\label{sec:pouct}
Despite our action abstraction to simplify planning, the space of collective actions can become large in practice, making calculating cost via Eq.~\eqref{eq:MR-LSP}  computationally intensive.
Instead, we rely on sample-based high-level planning and use a Partially Observable UCT (PO-UCT) \cite{silver2010monte}---a variant of Monte-Carlo tree search. PO-UCT is an anytime planning algorithm and allows us to approximate the expected cost without the need to exhaustively simulate all states.

During expansion of the planning search tree in PO-UCT, we maintain a \emph{rollout history} associated with each node: $\mathcal{H}_{b_t} = [[a_0,n_0,Q_0],\cdots,[a_k,n_k,Q_k]]$, that retains each node's history of (i) executed high-level actions $a_t$ and their outcomes, (ii) the number of times the node has been visited, and (iii) the accumulated expected cost $Q_t$ up to that node.
By simulating a collective action $a_t$ from the abstract state $b_t$, finding the subgoal $\sigma'$ via Eq.~\eqref{eq:frontiers-and-cost}, we expand the tree stochastically, where the outcome of a particular action is sampled from a Bernoulli distribution parametrized by the estimated $P_S(\sigma')$.
After the action outcome is sampled, the belief transitions to 
a node either corresponding to a success ($b_S$) or failure state ($b_F$), as defined in Eq.~\eqref{eq:updated-beliefs}.

Rollouts proceed similar to other Monte-Carlo Tree Search approaches.
The cost of each node corresponding to belief state ($b_S$ or $b_F$) is the sum of cost accrued to reach the current belief state from the initial belief state ($b_t$) and a search heuristic corresponding to the lower bound cost for the team to reach the goal if unseen space were assumed to be unoccupied.
After each node is visited, its count is incremented, which is used to control the rate of exploration during traversal. In all our \replaced{}{MR-LSP} experiments, we use 15,000 samples at each planning step.

\section{Training Data Generation and Learning Subgoal Properties} \label{sec:Training}
To compute the expected cost during MR-LSP planning, we require the subgoal properties $P_S$, $R_S$, and $R_E$ for all subgoals. We train a convolutional neural network, similar to that of \cite{bradley2021learning}, to estimate these properties from images collected by the robot.

The convolutional neural network (CNN) takes a $128\times 512$ RGB panoramic image aligned towards subgoal, egocentric position of the subgoal and goal as inputs. The image is passed through 4 convolutional layers, after which the relative-distance features are concatenated, passed through an additional 9 convolutional layers, and finally 5 fully connected layers, which output the subgoal properties. Our CNN estimates the subgoal properties for all subgoals and, with distances $D$ computed from the occupancy grid, are used to compute cost of a collective-action via Eq.~\eqref{eq:MR-LSP}.

To generate training data, we use a non-learned optimistic planner to navigate previously unseen environments. Observations (images) and  are collected from every step and labels of for each subgoal for $P_S$, $R_S$, and $R_E$ are computed from the underlying known map.
$P_S=1$ if the subgoal leads to the goal, for which $R_S$ is the distance to reach the goal, and $P_S=0$ if it does not, where $R_E$ is the distance the robot would travel before reaching a dead end and turning around.

\section{Experimental Results}

We conduct simulated experiments in two different environments---our own \textit{guided maze} and \textit{office floorplan} environments---in which the robot navigates from a randomly generated start location to a randomly generated point goal in unseen space. We evaluate the following approaches:
\begin{LaTeXdescription}
    \item[Multi-Robot Learning over Subgoals (MR-LSP)] Our approach, in which the expected cost of a collective action is computed via Eq.~\eqref{eq:MR-LSP}. When only one robot is used, this approach corresponds to LSP planning via Eq.~\eqref{eq:LSP}.
    \item[Non-Learned Optimistic Planner] We assume that all unseen space is unoccupied and compute paths through each subgoal to reach the goal. The optimistic plan cost is computed for each subgoal and linear sum assignment~\cite{Kuhn2010TheHM} is used to ensure that robots pursue different subgoals (a topological constraint on their plans similar to~\cite{bhattacharyaTopologicalConstraints}) while minimizing net optimistic plan cost to reach the goal. This planning strategy serves as a \emph{non-learned baseline}.
    \item[Linear Sum Assignment using LSP (LSA-LSP)] This planning strategy also uses linear sum assignment to enforce that robots select different subgoals to pursue, yet the expected cost associated with each subgoal is computed via the single-robot LSP approach Eq.~\eqref{eq:LSP}.
    This planning strategy serves as a \emph{learning-informed baseline}. When one robot is used, this approach corresponds to LSP planning via Eq.~\eqref{eq:LSP}.
    \item[Known-Space Planner] Planning in the fully-known map (no uncertainty), providing a lower bound on possible cost. Since the shortest path is known, this planner has the same performance for any number of robots.
\end{LaTeXdescription}

\begin{figure}[t]
    \vspace{1em}
    \centering
    \footnotesize
    \begin{tabular}{cccc}
    \toprule
     & \multicolumn{3}{c}{\textbf{Guided Maze Environment}}           \\
    \cmidrule(lr){2-4}
             Planner                & 1 robot         & 2 robots        & 3 robots       \\
    \midrule
    Non Learned Optimistic   & 207.50          & 144.88          & 144.44         \\
    LSP-LSA (Learned)        & \textbf{162.46}          & \textbf{133.12}          & \underline{132.66}         \\
    MR-LSP (Ours)            & \textbf{162.46} & \underline{134.22} & \textbf{132.3} \\
    \midrule
    Known-Map Planner            & 130.55          & 130.55          & 130.55        \\
    \bottomrule
    \\[-6pt]
    \end{tabular}
    \includegraphics[width=8.45cm]{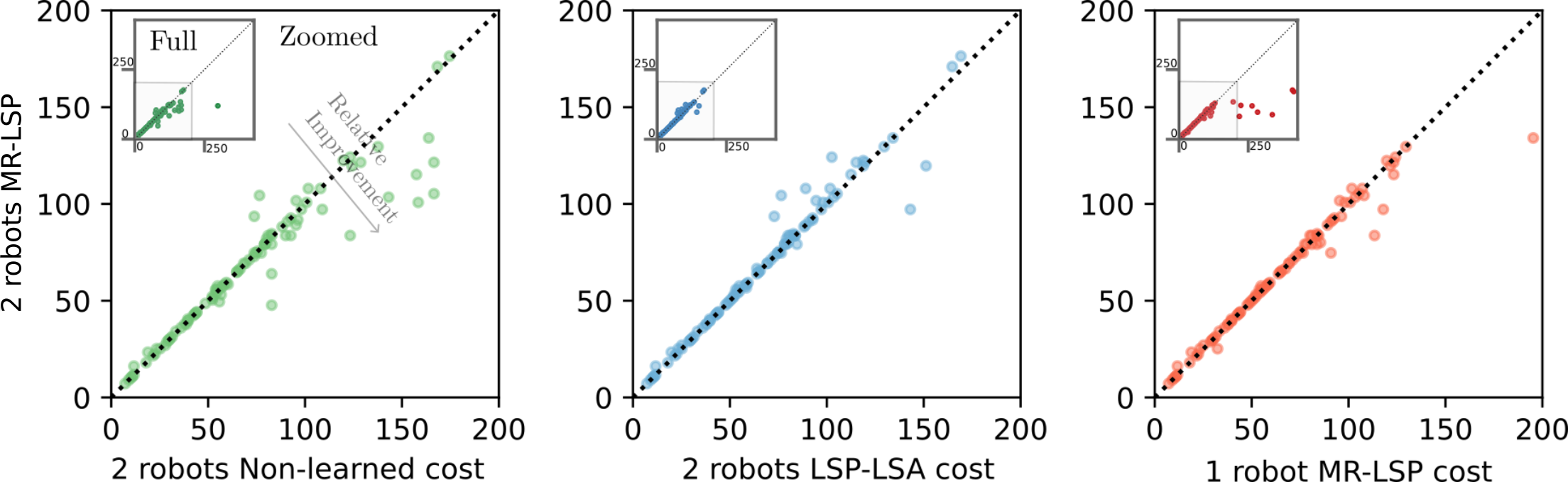}
    \caption{\textbf{Navigation in maze environment results}\quad The table shows the average cost (in meters) accrued in 100 experiments for each planner. The scatter plot each shows the performance of our approach versus the baselines for two robots. The result shows planners which use learning (MR-LSP and LSP-LSA) outperform the non-learned planner. The performance of learned baseline (LSP-LSA) is similar to MR-LSP (slightly better for two robots) in a relatively simpler environment.}   
    \vspace{-1.5em}\label{fig:scatter_result_maze}
\end{figure}

\subsection{Guided Maze Environment Results}
We first perform experiments in our ``guided'' maze environment, in which a green path on the ground connects the goal to another point in the maze. Initially, the robot team is placed at the center of this path so that two green routes extend from its location, only one leading to the goal.
The maze is simply connected---i.e., there exists only a single path to the goal---and so a single robot, even one that understands the significance of the green path, must get lucky to reach the goal quickly by choosing the correct path. However, a two robot team can divide-and-conquer, following both green paths simultaneously and reaching the goal quickly with high reliability.

\begin{figure}[t]
        \vspace{1em}
        \centering
        \includegraphics[width=8.45cm]{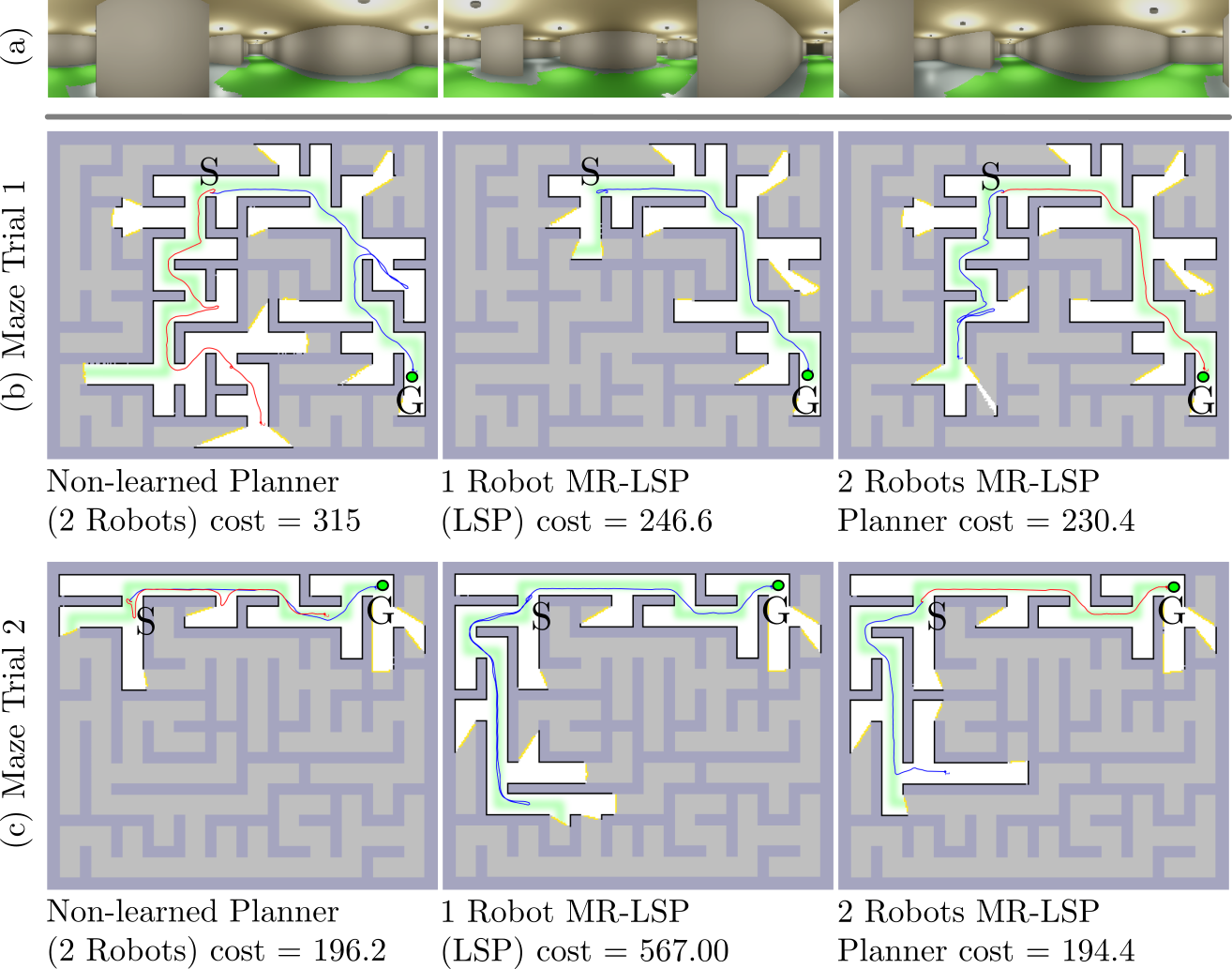}
        \caption{\textbf{Navigation under uncertainty in maze} \quad (a) Images taken on board robot used as inputs to learning. (b) Generally learned planner knows to follow green path to reach goal quickly. (c) Occasionally, MR-LSP with single robot (LSP) explores alternative route, whereas with multiple robots MR-LSP guides coordinated exploration.}
        \label{fig:MazeResult}
        \vspace{-1.5em}
\end{figure}

We evaluate \replaced{our planners}{each of our planner} in 100 guided maze environments, and show that our \replaced{MR-LSP}{Multi-Robot Learning over Subgoals Planner (MR-LSP)} and the LSA-LSP learned baseline perform near-optimally in this environment for 2- and 3-robot experiments, outperforming both the 1-robot trials and the non-learned optimistic baseline. Fig.~\ref{fig:scatter_result_maze} shows the average cost for each planner in this environment.
\replaced{Both MR-LSP and LSA-LSP planners use learning to evaluate goodness of paths and understand the importance of splitting the team to follow both green paths for reaching the goal. In contrast, the non-learned optimistic planner explores many unlikely paths due to a lack of understanding of the green path's significance. These behaviors are evident in Fig.~\ref{fig:MazeResult}(b) and (c), highlighting the significance of using learning to prioritize the green path and follow multiple promising routes simultaneously.}{Both the MR-LSP and LSA-LSP planners rely on learning to evaluate the goodness of paths to the goal and correctly understand that the team should split up and follow both green paths simultaneously to reach the goal.
By contrast, the non-learned optimistic planner does not understand the significance of the green path and explores many paths unlikely to lead to the goal. These behaviors can be seen in Fig.~\ref{fig:MazeResult}(b) and (c), demonstrating both the importance of using learning to prefer the green path and of splitting up to follow multiple promising routes simultaneously.}

\subsection{Office Floorplan Environment Results}
\begin{figure}[t]
    \vspace{1em}
    \centering
    \footnotesize
    \begin{tabular}{cccc}
    \toprule
     & \multicolumn{3}{c}{\textbf{Office Floorplan Environment}}    \\
    \cmidrule(lr){2-4}
                   Planner          & 1 robot         & 2 robots      & 3 robots       \\
    \midrule
    Non Learned Optimistic   & 206.90          & 167.56        & \underline{156.25}         \\
    LSP-LSA (Learned)        & \textbf{159.84}          & \underline{161.96}        & 158.23         \\
    MR-LSP (Ours)            & \textbf{159.84} & \textbf{145.28} & \textbf{149.10} \\
    \midrule
    Known-Map Planner            & 129.48          & 129.48        & 129.48         \\
    \bottomrule
    \\[-6pt]
    \end{tabular}
    \includegraphics[width=8.45cm]{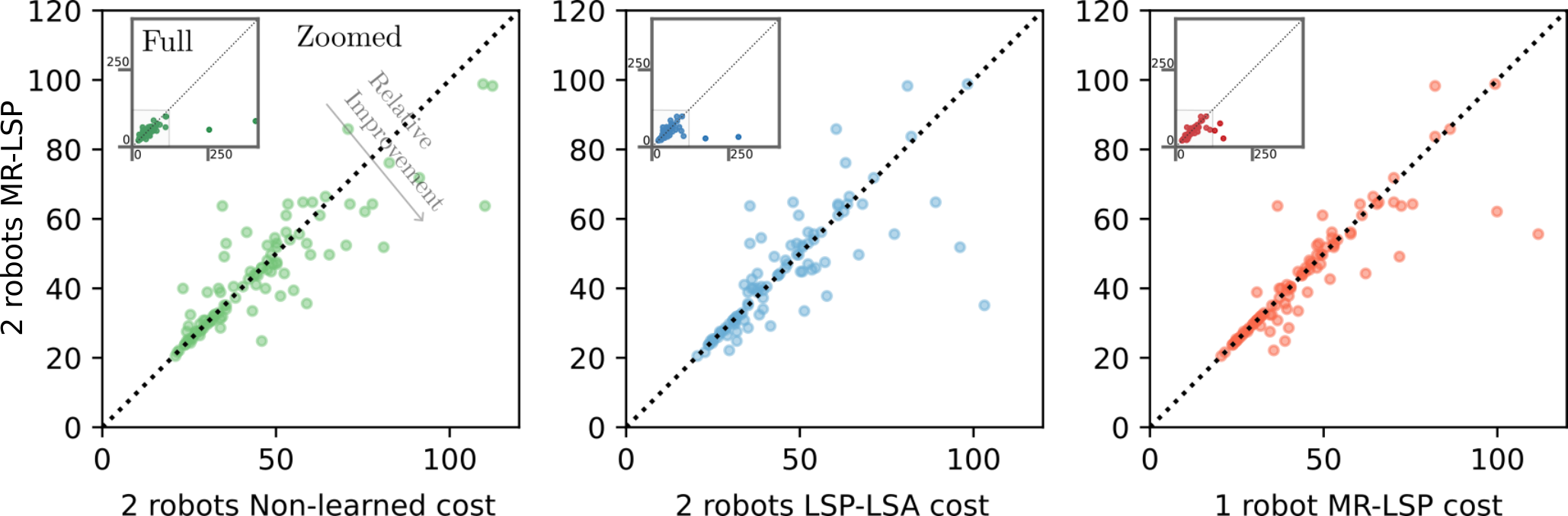}
    \caption{\textbf{Navigation in office floorplan results} \quad The table shows the average cost (in meters) accrued in 100 experiments for each planners. The scatter plot each shows performance of our approach versus the baselines for two robots. In a simulated office floorplan, MR-LSP outperforms both non-learned and learned baselines.   }
    \vspace{-15pt}
    \label{fig:scatter_result_office}
\end{figure}

\replaced{We conduct experiments in our simulated \emph{office floorplan} environments. These environments are designed to mimic a typical office building, with multiple hallways intersecting and connecting many offices, including some corner rooms. Clutter in each office simulates furniture and obstructs view from the hallway. We evaluate our planners in 100 procedurally generated office environments---distinct from those seen during training---with varying numbers of robots and report the effectiveness of our MR-LSP planning approach in Fig.~\ref{fig:scatter_result_office}.}{ We additionally conduct experiments in our simulated \emph{office floorplan} environments.
These environments are designed so as to mimic a typical office building, with multiple intersecting hallways each with many offices connected to them, including some corner rooms connected to two hallways.
Clutter in each office simulates furniture and obstructs view of the entire office from the hallway.
We evaluate in 100 office environments for each of our planners and number of robots and report results in Fig.~\ref{fig:scatter_result_office} that demonstrate the effectiveness of our MR-LSP planning approach.}

\begin{figure}[t]
        \vspace{1em}
        \centering
        \includegraphics[width=8.45cm]{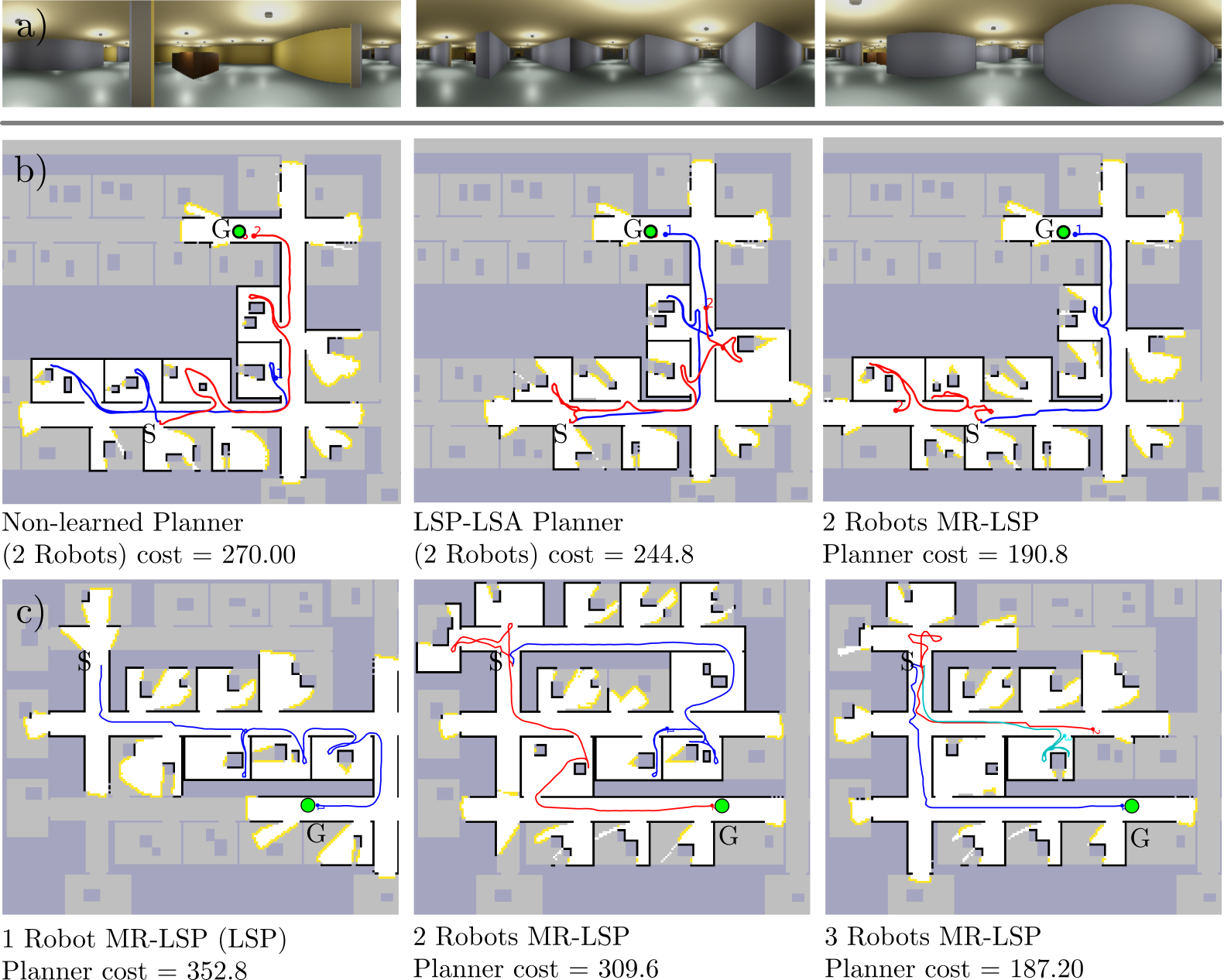}
        \vspace{-0.4em}
        \caption{\textbf{Navigation under uncertainty with two robots in the simulated office floorplan} \quad (a) Images from office used as inputs to learning (b) Learned planner improves cost compared to the non-learned planner. MR-LSP performs better over non-learned and learned baselines. (c) MR-LSP generally improves cost as the number of robots increase.}
        \label{fig:office1}
\end{figure}

Our MR-LSP approach seems to both understand the utility of following hallways until the goal can be reached and how to effectively allocate exploratory actions to different team members, improving performance over both LSA-LSP and Non-Learned Optimistic planners.
\replaced{The LSA-LSP learned baseline, which uses the same subgoal estimator as the MR-LSP planner, tends to follow hallways and avoid exploring rooms. However, due to a lack of coordination among the team, it only moderately improves over the non-learned baseline. Fig.~\ref{fig:office1}b shows an example of this behavior. The LSA-LSP planner can only coordinate two robots myopically, resulting in poorer behavior compared to the MR-LSP planner, which quickly explores promising routes to the goal. }{The LSA-LSP learning-informed baseline, which uses the same subgoal property estimator as does the MR-LSP planner, also tends to follow hallways, yet the lack of non-myopic coordination amongst the team, leading to only moderate improvements over the non-learned baseline.

We show a qualitative example of this behavior in Fig.~\ref{fig:office1}b. The LSA-LSP (learned baseline) planner knows to follow hallways and avoid exploring rooms, unlike the non-learned planner. However, the LSA-LSP planner can only myopically coordinate two robots, resulting in poor behavior when compared to the MR-LSP planner, which more quickly explores the promising routes to the goal.}

\begin{figure}[t]
        \centering
        \includegraphics[width=8.45cm]{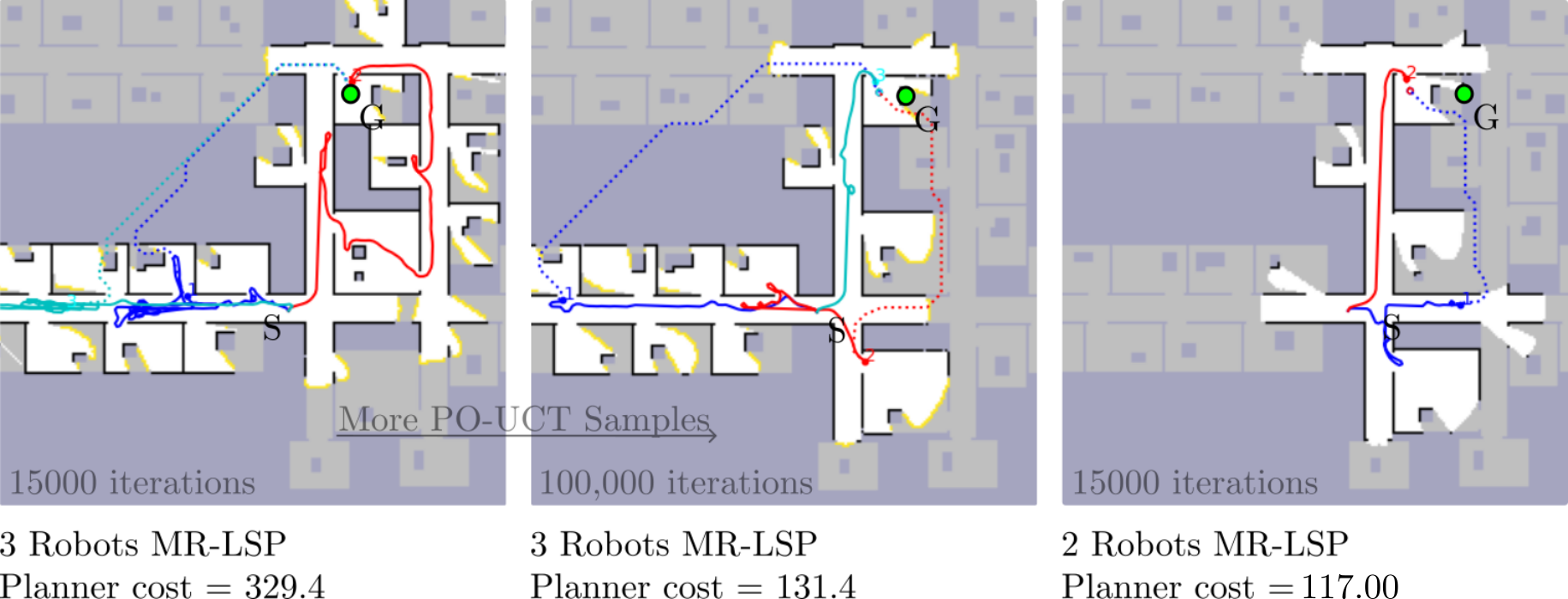}
        \vspace{-0.4em}
        \caption{Increasing the number of PO-UCT samples improves performance for three robots.}
        \label{fig:3robots}
\end{figure}

\replaced{In the office floorplan environments, 3-robot MR-LSP experiments show \emph{increased} average cost compared to 2-robot MR-LSP experiments in the same space. Fig.~\ref{fig:3robots} shows a scenario where 2-robot MR-LSP outperforms 3-robot MR-LSP planning. Additional experiments increasing the number of samples for PO-UCT from 15k to 100k showed improved performance, indicating that the slight degradation is owed to a limited computational budget and should not be seen as a limitation of the approach itself.}{We point out that average cost \emph{increases} for the 3-robot MR-LSP experiments in the office floorplan environments compared to 2-robot MR-LSP experiments in the same space. Fig.~\ref{fig:3robots} shows an example of a scenario in which 2-robot MR-LSP outperforms the 3-robot MR-LSP planning. We conducted additional experiments on this and other trials in which we increased the number of samples for PO-UCT from 15k to 100k and found that the performance improved suggesting that the slight performance degradation is owed to a limited computational budget and should not be seen as a limitation of the approach itself.}
In future work, we will explore this relationship in more depth and reimplement our planner in a faster, compiled language.

\section{Related Works}
Multi-robot navigation in unknown environment is often modelled as a Decentralized POMDP (Dec-POMDP)~\cite{DECPOMDP_MAIN1,DECPOMDPMain2,DecPOMDPintractable}.
Owing to computational challenges in Dec-POMDP planning, dynamic programming~\cite{hansen2004dynamic} and heuristic search approaches~\cite{szer2012maa} are limited to comparatively short-horizon planning tasks.

\textbf{Learning for Multi-Robot Planning}\quad{}
Learning is frequently used to address the difficulties of planning under uncertainty. Many learning-informed approaches in this domain (e.g.,~\cite{gupta,richter1}) focus on single-robot planning and do not scale to larger and more complex environments. Multi-agent reinforcement learning has a long history in this domain \cite{lauer2000algorithm,matignon2012independent}, yet only with recent advances in deep reinforcement learning has it become possible to navigate in somewhat realistic environments \cite{marl, rl1, rl2, RLsurvey}. Still, though these approaches work well for small environments, they are often brittle to change~\cite{henderson2018rlmatters} and struggle to scale to large-scale environments at the scale of buildings.

\textbf{Action Abstraction for Multi-Robot Planning}\quad{}
To mitigate computational challenges, many approaches in this domain rely on a state or action abstraction to simplify planning. 
Temporally-extended \textit{macro-actions} are one such action abstraction that help to scale planning under uncertainty~\cite{MACROACTION1, MACROACTION2, MACROACTION3}. However, these approaches have not proven scalable to building sized environments. Some strategies~\cite{hoerger2019multilevel, amato2015scalable} use variants of Monte Carlo tree search for faster computation in POMDP for long-horizon planning, yet without direct access to a distribution over environments are limited in their ability to reason far into the future.
Other approaches to multi-robot planning introduce \emph{topological action abstractions} to simplify planning~\cite{frontier-multirobot,topological-mr,frontier-mr-1}. Under these approaches, each robot is constrained to leave known space through different \textit{frontiers}, boundaries between known space and unknown space \cite{frontier-singlerobot}, or via paths belonging to different relative homology~\cite{BhattacharyaHomologyPathPlanning, bhattacharyaTopologicalConstraints}. Works on navigation using this abstraction generally uses greedy heuristic to select where the robot should navigate next and do not take into account the impact of an action over the longer horizon. 

The Learning over Subgoals Planning (LSP) approach~\cite{stein2018learning} uses both a model-based topological abstraction and supervised learning to improve navigation performance, yet is designed only with a single robot planning in mind.


\section{Conclusion}
In this work, we present Multi-Robot Learning over Subgoals (MR-LSP): a novel method for performant and reliable, learning-informed multi-robot navigation through partially-mapped environments.
Using learning to estimate the goodness of individual exploratory actions that enter unseen space, our multi-robot team is able to envision the expected long-horizon impact of each robot's actions and can thus coordinate behavior far into the future to quickly reach the unseen goal, outperforming both learned and non-learned strategies. However, the number of multi-robot actions grows rapidly with the number of robots, imposing a practical limitation on team size.
In future work, we would like to extend our model to support more complex and multi-stage tasks, effectively extending the model of Bradley et al.~\cite{bradley2021learning} to the multi-robot planning domain.


\section*{Acknowledgement}
We thank Kevin Doherty, Chris Bradley, Jana Kosecka, Erion Plaku, and Cameron Nowzari for their thoughtful feedback on this work. We acknowledge funding support from George Mason University.

\bibliography{references.bib}

\end{document}